\pdfoutput=1

\documentclass[11pt]{article}

\usepackage[]{acl}

\usepackage{times}
\usepackage{latexsym}

\usepackage[T1]{fontenc}

\usepackage[utf8]{inputenc}

\usepackage{microtype}

\usepackage{amsmath,amsfonts,bm}

\def\eqref#1{equation~\ref{#1}}

\def\1{\bm{1}}

\DeclareMathAlphabet{\mathsfit}{\encodingdefault}{\sfdefault}{m}{sl}
\SetMathAlphabet{\mathsfit}{bold}{\encodingdefault}{\sfdefault}{bx}{n}

\usepackage{url}
\usepackage{multirow}
\usepackage{booktabs}
\usepackage{graphicx}

\title{Meta-learning via Language Model In-context Tuning}

\author{
Yanda Chen\textsuperscript{1}\thanks{\hspace{2mm}Work done during summer internship at AWS AI.}~~~~~~Ruiqi Zhong\textsuperscript{2}~~~~~~Sheng Zha\textsuperscript{3}~~~~~~George Karypis\textsuperscript{3}~~~~~~He He\textsuperscript{34}\\
\textsuperscript{1}Columbia University,
\textsuperscript{2}University of California, Berkeley,
\textsuperscript{3}AWS AI\\
\textsuperscript{4}New York University\\
{\tt yc3384@columbia.edu, ruiqi-zhong@berkeley.edu,} \\
{\tt \{zhasheng, gkarypis, hehea\}@amazon.com}\\
}

\begin{document}
\maketitle

\begin{abstract}
The goal of meta-learning is to learn to adapt to a new task with only a few labeled examples. Inspired by the recent progress in large language models, we propose \textit{in-context tuning} (ICT), which recasts task adaptation and prediction as a simple sequence prediction problem: to form the input sequence, we concatenate the task instruction, labeled in-context examples, and the target input to predict; to meta-train the model to learn from in-context examples, we fine-tune a pre-trained language model (LM) to predict the target label given the input sequence on a collection of tasks.

We benchmark our method on two collections of text classification tasks: LAMA and BinaryClfs. Compared to MAML which adapts the model through gradient descent, our method leverages the inductive bias of pre-trained LMs to perform pattern matching, and outperforms MAML by an absolute 6\% average AUC-ROC score on BinaryClfs, gaining more advantage with increasing model size. Compared to non-fine-tuned in-context learning (i.e.\ prompting a raw LM), in-context tuning meta-trains the model to learn from in-context examples. On BinaryClfs, ICT improves the average AUC-ROC score by an absolute 10\%, and reduces the variance due to example ordering by 6x and example choices by 2x. \footnote{Code is released at \url{ https://github.com/yandachen/In-context-Tuning}.}

\end{abstract}

\section{Introduction}
Few-shot learning (FSL) refers to a system's ability to quickly adapt to new tasks when very few labeled examples are available for training. 
FSL is a key feature of human learning \cite{lake2016building}, but current machine learning systems often rely on large amounts of labeled training data \cite{alphago, 7780459, adiwardana2020humanlike}.

\begin{figure*}
    \centering
    \includegraphics[width=\linewidth]{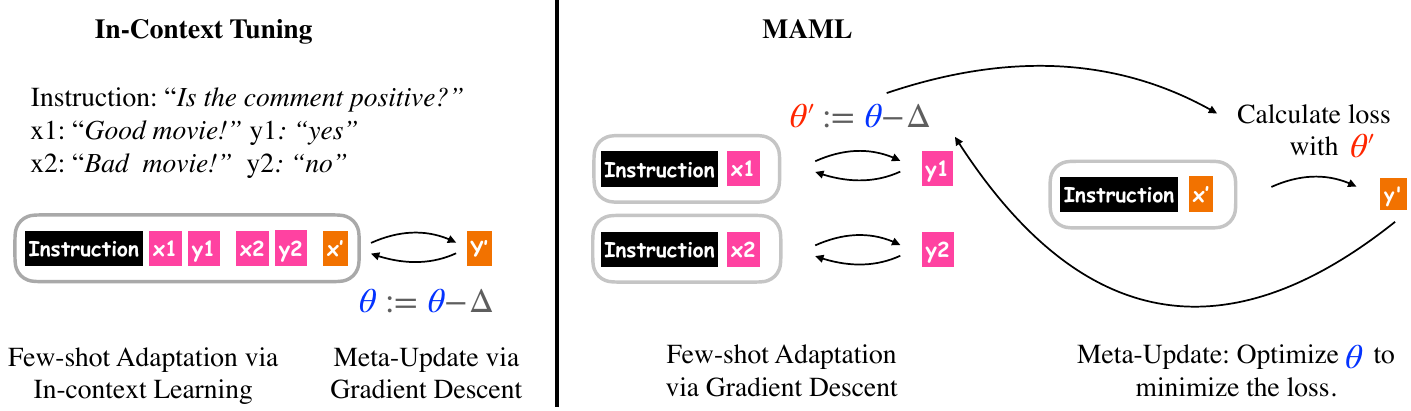}
    \caption{\textbf{MAML} (right): MAML aims to learn a task-agnostic model initialization $\theta$ that can adapt fast to new tasks. To adapt the model initialization to a new task $\Tilde{T}$, a task-specific model $\theta'$ initialized with $\theta$ is updated with gradient descent using task examples from $\Tilde{T}$. Meta-training of MAML involves bi-level optimization, where the inner optimization learns a task-specific model $\theta'$ using task examples from $\Tilde{T}$, and the outer optimization learns a meta-initialization $\theta$ to minimize few-shot prediction loss of $\theta'$ on task $\Tilde{T}$. \textbf{In-context Tuning (ours)} (left): our approach adapts to new tasks via in-context learning, and learns a single model $\theta$ shared across all tasks that is directly optimized with the FSL objective (Section~\ref{sec:ouralgo}). Because model parameters are frozen during task adaptation, our approach does not involve bi-level optimization during meta-training. 
    }
    \label{fig:maml-ICT-comparison}
\end{figure*}

Recently, prompting large pre-trained language models (LMs) for FSL has achieved remarkable progress \cite{brown2020language, schick-schutze-2021-exploiting}.
LM prompting with in-context learning reduces the ``task learning and predict'' process to a simple sequence prediction problem.
To perform a new task, \citet{brown2020language} prompt a \textit{raw} LM (i.e., a pre-trained LM not fine-tuned on any labeled data) with the concatenation of the task instruction, some input-output examples, and the target input to be predicted on; then they extract the answer from the LM's continuation of the concatenated sequence (Figure~\ref{fig:maml-ICT-comparison} left). 
For example, to coax the model into performing sentiment classification on the target input ``\textit{This movie is a waste of time}'', we prompt the LM with the sequence ``\textit{I like the movie! Positive review? Yes. Horrible Movie! Positive review? No. This movie is a waste of time. Positive review? \_\_\_}'', and predict ``positive'' if the next word is more likely to be ``\textit{Yes}'' rather than  ``\textit{No}''. 

However, raw LMs are not optimized for in-context FSL during pre-training, and exhibit undesirable behavior when used for FSL. 
For example, \citet{zhao2021calibrate} observed that LMs suffer from the ``recency bias'', which assigns higher probability to labels that appear closer to the target input. 
As a result, the accuracy becomes extremely sensitive to the ordering of the in-context examples. 
Previous work has also shown that prompting raw LMs is often oversensitive to example choices and instruction wording \cite{schick-schutze-2021-exploiting, 10.1162/tacl_a_00324, gao2021making, liu2021makes}.

We address this weakness through a meta-learning lens and directly fine-tune the LM for FSL.
Under the meta-learning framework, we meta-train a model to learn to adapt to new tasks from a few examples on a wide range of tasks, so that it learns to leverage the few-shot examples to adapt to new tasks at test time. 
Since LM prompting already reduces the ``task learning and predict'' process
to a simple sequence prediction problem, we 
meta-train a LM by directly fine-tuning it to optimize for this sequence prediction problem on a wide range of tasks \mbox{(Figure \ref{fig:maml-ICT-comparison} left)}.
Since we fine-tune our model to learn in-context learning, we call our approach \textit{\textbf{in-context tuning}} (ICT).
Unlike optimization-based meta learning approaches such as MAML \cite{finn2017modelagnostic},
in-context tuning adapts to new tasks through in-context learning where model parameters are frozen, thus it avoids the challenging nested optimization problem in MAML (Figure \ref{fig:maml-ICT-comparison}).

We benchmark our algorithm on LAMA \cite{petroni-etal-2019-language}, a dataset for testing models' factual knowledge, and BinaryClfs \cite{zhong2021adapting}, a wide range of binary classification tasks each annotated with a few language descriptions of the task.
Compared to prompting raw LMs, in-context tuning improves performance by 7.6 Precision@1 points on LAMA and 10.6\% AUC-ROC score on BinaryClfs. 
In addition, in-context tuning mitigates the over-sensitivity of raw LM prompting, significantly reducing the variance of the performance with respect to example ordering (by 68\% on LAMA and 83\% on BinaryClfs), example choices (by 56\% on LAMA and 40\% on BinaryClfs), and instruction wording (by 19\% on LAMA).

Our approach also out-performs MAML, which adapts the model by gradient descent on a few examples and learns an initialization that can adapt to a new task through a few gradient steps \cite{finn2017modelagnostic, nichol2018firstorder}.
Since our approach better takes advantage of the inductive bias of LMs to extrapolate from in-context examples, our approach out-performs first-order MAML by 2.8 points on LAMA and 5.1 points on BinaryClfs, with increasing advantage as models become larger.

Given the empirical effectiveness of in-context tuning (Section~\ref{sec:accuracy}), we conjecture that the few-shot learning potential of large LMs (e.g., \mbox{GPT-3}) may be broadly underestimated if prompted without any direct optimization for FSL. We also conjecture that in-context tuning can mitigate various undesirable properties of LM prompting, such as over-sensitivity to example ordering, example choices, and instruction wording \mbox{(Section~\ref{sec:sensitivity})}.

\section{Approach}

We introduce the problem setup (Section~\ref{sec:setup}), describe our in-context tuning algorithm (Section~\ref{sec:ouralgo}), compare our algorithm to gradient-based adaptation methods (Section~\ref{sec:gradient-based-adapt}) and other baselines (Section~\ref{sec:other-baselines}). 

\subsection{Problem Setup} \label{sec:setup}
We focus on the few-shot classification problem, where the model first learns from a set of training tasks $T \in T_{\text{train}}$, each associated with its natural language instructions $I_{T}$ and a large amount of task input-output examples $D_T = \{(x^{i}_{T}, y^{i}_{T})\}$ (see Figure~\ref{fig:maml-ICT-comparison} left for examples). 
At test time, we ask the model to learn a new task $\Tilde{T}$ given its instruction and only a few ($K$) labeled examples, i.e. $S_{\Tilde{T}} \subseteq D_{\Tilde{T}}, |S_{\Tilde{T}} | = K $. 
We denote the task input to be predicted at test time as $x_{\Tilde{T}}^{\text{target}}$.

Note that ``task input'' is different from ``model input''.
For example, on the left panel of Figure \ref{fig:maml-ICT-comparison}, the task input is ``\textit{Good movie!}'' while the model input can be a concatenation of the instruction, task inputs and task outputs.

\subsection{In-context Tuning Algorithm} \label{sec:ouralgo}
In-context tuning directly optimizes pre-trained LMs with the few-shot in-context learning objective \cite{brown2020language}: task-agnostic LMs are meta-trained to perform few-shot in-context learning on a wide variety of training tasks. 
Similar to in-context learning, LMs trained with in-context tuning adapt to a new task by using few-shot training examples as the input prefix. 

Formally, during meta-training, we build the model input by concatenating the task instruction $I_{T}$, task input-output pairs $S_{T}  \subseteq D_{T}$, and the task input $x^{\text{target}}_{T}$\footnote{We sometimes abbreviate ``target'' as ``tgt'' to save space.} to be classified. 
We then fine-tune a pre-trained LM to predict $y^{\text{target}}_{T}$ and hope that the model learns to use the in-context examples $S_{T}$. Here is the few-shot in-context tuning objective $\mathcal{L}$: 

\begin{align}
\mathcal{L}_{T}(\theta) &:= \sum_{ (x^{\text{tgt}}_T, y^{\text{tgt}}_T) \in D_T}[-\log p_{\theta}(y^{\text{tgt}}_T | x^{\text{tgt}}_T, S_T, I_{T})]\\
\mathcal{L}(\theta) &:= \sum_{T \in T_{\text{train}}} \mathcal{L}_{T}(\theta)
\end{align}

To adapt to a new task $\Tilde{T}$ at test time, we directly concatenate the few-shot examples $S_{\Tilde{T}}$ with the instruction $I_{\Tilde{T}}$ and the target task input $x^{\text{target}}_{\Tilde{T}}$ to be classified to form the model input, and ask the model to predict its corresponding output. 
No gradient update is performed during adaptation.

\subsection{Gradient-based Task Adaptation}
\label{sec:gradient-based-adapt}
We compare in-context tuning with two classical few-shot learning methods: multi-task fine-tuning (instruction tuning + fine-tuning) and MAML.
Both methods adapt the model parameters to new tasks by gradient descent on few-shot examples.

\paragraph{Instruction Tuning + Fine-tuning (InsT + FT)}
We extend the recent work on zero-shot instruction tuning \cite{wei2021finetuned} to the FSL setting as a \textit{multi-task fine-tuning} baseline.
During meta-training, the model is optimized to predict the task output given the task instruction and the task input on a wide range of tasks \cite{zhong2021adapting}. 
Formally, we train the model parameter $\theta$ to predict $y^{i}_{T}$ given $I_{T} \circ x^{i}_{T}$, where $\theta$ is shared across all tasks and $\circ$ represents the concatenation operation.
During the few-shot adaptation phase, the model is presented with a new task $\Tilde{T}$, its natural language instruction $I_{\Tilde{T}}$ and a small set of ($K$) task input-output examples $S_{\Tilde{T}} = \{(x^{i}_{\Tilde{T}}, y^{i}_{\Tilde{T}}) | i \in [K]\}$. 
We then fine-tune the model to predict the task output $y^{i}_{\Tilde{T}}$ from the new task given $I_{\Tilde{T}} \circ x^{i}_{\Tilde{T}}$ and update $\theta$ with a few gradient steps to get $\theta_{\Tilde{T}}$.
Finally, we use the updated model $\theta_{\Tilde{T}}$ to predict the output from the task input $x^{\text{target}}_{\Tilde{T}}$ and the instruction $I_{\Tilde{T}}$ under the test task $\Tilde{T}$.

\paragraph{MAML}
The few-shot adaptation stage of MAML is the same as instruction tuning + fine-tuning, where we update the model parameters (initialized with $\theta$) by gradient descent on $K$ examples $S_{\Tilde{T}} \subseteq D_{\Tilde{T}}$.
However, during meta-training, MAML aims to learn a task-agnostic model initialization $\theta$ such that, $\theta_{T}$, which is to be found by initializing with $\theta$ and  performing gradient descent on $S_{T}$, would lead to good performance \cite{finn2017modelagnostic}.

Therefore, MAML involves two levels of optimization, an inner optimization to learn $\theta_T$ given $\theta$ and $S_T \subseteq D_T$, and an outer optimization to learn $\theta$ given $\theta_T$. 
Due to the bi-level structure in this optimization problem, MAML has been found to be empirically unstable, sensitive to hyperparameters, and computationally expensive \cite{finn2017modelagnostic, nikolaev-etal-2020-fine}. 
Even worse, few-shot task adaptation is known to be highly sensitive to optimization hyperparameters \cite{antoniou2018how}, while a large labeled validation set for hyperparameter tuning may not be available under a FSL setting \cite{perez2021true}. 

In comparison, in-context tuning simplifies the two-stage process of (1) few-shot task adaptation and (2) task-specific prediction as one sequence prediction problem, where task-specific examples are concatenated to the model input to provide information about the task.
Hence, in-context tuning removes the bi-level optimization during meta-training, which can be empirically unstable and expensive. 
Additionally, since model weights are frozen during task adaptation, it is not sensitive to adaptation hyperparameters.

\subsection{Other Baselines}
\label{sec:other-baselines}
\paragraph{Raw In-context Learning (Raw IC-L)} 
We directly evaluate a raw LM on a new task using the same evaluation set-up for in-context tuning, without fine-tuning the LM on any labeled data. 

\paragraph{Instruction Tuning (InsT)} 
The model learns to predict the target output only based on the instruction and the target input. 
Only the instruction is available during the adaptation phase, and this setup is also known as zero-shot learning.

We categorize all approaches in our paper based on their meta-training objective and how they use task-specific examples in Table \ref{tab:methodinfo}. In-context tuning is the \textit{only} method that directly optimizes the FSL objective without gradient-based adaptation.

\begin{table}[t]
    \centering
    \begin{tabular}{lcc}
        Method & Adaptation & Meta-train  \\
        \hline 
        In-context Tuning & In-context & Few-shot \\
        MAML & Gradient & Few-shot \\
        InsT & None & Zero-shot \\
        InsT + FT & Gradient & Zero-shot\\
        Raw IC-L & In-context & LM\\
        \hline
    \end{tabular}
    \caption{We categorize our approach and the baselines according to 1) how the few-shot examples (if any) are used for adaptation, and 2) the meta-training objective. Ins-T refers to instruction tuning.}
    \label{tab:methodinfo}
\end{table}


\begin{table*}[t]
\small
\setlength{\tabcolsep}{4pt}
\centering
\begin{tabular}{lcccccccccccccccc}
\toprule
\multirow{3}{*}{} & \multicolumn{12}{c}{\textbf{LAMA}} & \multicolumn{4}{c}{\textbf{BinaryClfs}} \\
\cmidrule(lr){2-13} \cmidrule(lr){14-17}
& \multicolumn{4}{c}{\textbf{BERT-Base}} & \multicolumn{4}{c}{\textbf{BERT-Large}} & \multicolumn{4}{c}{\textbf{DeBERTa-xlarge}} & \multicolumn{2}{c}{\textbf{GPT2-M}} & \multicolumn{2}{c}{\textbf{GPT2-L}}\\
\cmidrule(lr){2-5} \cmidrule(lr){6-9} \cmidrule(lr){10-13} \cmidrule(lr){14-15} \cmidrule(lr){16-17}
& 0-S & 1-S & 2-S & 5-S & 0-S & 1-S & 2-S & 5-S & 0-S & 1-S & 2-S & 5-S & 0-S & 5-S & 0-S & 5-S \\
\cmidrule{1-17}
Raw IC-L & 10.3 & {8.5} & {10.8} & {14.1} & 12.7 & {12.1} & {15.4} & {18.6} & 11.2 & 12.6 & 20.6 & 23.7 & 50.5 & 57.8 & 51.0 & 58.3 \\
\cmidrule{1-17}
InsT + FT & / & \textbf{17.5} & \textbf{18.6} & \textbf{20.0} & / & \textbf{21.6} & 22.6 & 23.9 & / & 24.7 & 25.6 & 27.0 & / & 67.0 & / & 69.4 \\
\cmidrule{1-17}
ICT & \textbf{14.6} & 16.3 & 17.6 & 19.6 & \textbf{18.0} & \textbf{21.6} & \textbf{23.4} & \textbf{24.3} & \textbf{21.9} & \textbf{26.0} & \textbf{27.5} & \textbf{28.8} & \textbf{62.9} & \textbf{67.4} & \textbf{66.3} & \textbf{69.8} \\
\midrule
Raw IC-L w/o Ins & 1.5 & 4.9 & 8.7 & 12.3 & 1.4 & 3.5 & 7.0 & 12.5 & 2.7 & 13.0 & 19.5 & 22.6 & / & / & / & / \\
\cmidrule{1-17}
ICT w/o Ins & 7.1 & 14.6 & 17.0 & 18.2 & 9.3 & 19.4 & 19.9 & 22.9 & 10.6 & 23.5 & 26.0 & 27.6 & / & / & / & / \\
\bottomrule
\end{tabular}
\caption{\label{tab:ict-vs-baselines} Few-shot learning accuracy of our in-context tuning approach (ICT) compared to in-context learning with raw LMs (Raw IC-L) and instruction tuning + fine-tuning (InsT + FT). $K$-S: $K$-shot learning. GPT2-M: GPT2-Medium. GPT2-L: GPT2-Large. Task instructions are used except the last two rows labeled with ``w/o Ins''. By definition, InsT + FT is the same as ICT for 0-S. We only experiment with the no-instruction setting on the LAMA dataset. Since we modify the LAMA dataset and BinaryClfs dataset (Section~\ref{sec:dataset-and-metrics}), the numbers reported in our work are not directly comparable to other work.}
\end{table*}

\section{Experimental Setup}


\subsection{Datasets and Metrics} \label{sec:dataset-and-metrics}

We experiment with two meta-datasets
that contain a wide range of tasks, LAMA and BinaryClfs. 
Each task is associated with several different natural language descriptions, and we call them \textit{instructions} for convenience, even though some of them are realized as questions.

\paragraph{LAMA} \textbf{LA}nguage \textbf{M}odel \textbf{A}nalysis \cite{petroni-etal-2019-language} is a dataset that tests the factual and commonsense knowledge learned by LMs. In our experiments, we use the TREx-UHN portion of LAMA \cite{poerner-etal-2020-e}, which consists of (subject, relation, object) triples from Wikidata. 
LAMA is an entity prediction task, where a model is asked to predict the object entity given the subject entity and the relation. In our experiments, we treat one relation as a task as in \citet{perez2021true}. 

Initial experiments on LAMA showed that LMs take significant advantage of ``majority label bias'' \cite{zhao2021calibrate}, where they assign higher probability to object entities that have appeared in the in-context examples, thus inflating the accuracy. 
To reflect the improvement due to few-shot learning rather than this simple heuristic to copy answers, for all tasks we prune the LAMA dataset so that all object entities appear less than 2.5\% of times. 
Our final filtered LAMA dataset consists of 29 relations (tasks) and 12k (subject, relation, object) examples. 

We use task instructions from two datasets: LAMA and LPAQA \cite{10.1162/tacl_a_00324}. 
LAMA contains one task instruction for each task, and the auxiliary LPAQA dataset contains on average 10 additional instructions for each LAMA task.

We use the same evaluation protocol as in \citet{petroni-etal-2019-language}: 1) the object entity is predicted from a pre-defined vocabulary set of 21k words (each LAMA task is 21k-way classification); 2) we compute mean precision at one (P@1) for each task, and report the average across tasks.
Because LAMA does not have an official train-validation-test split, we use 8-fold cross-validation in our experiments. We randomly partition the 29 tasks into 8 groups of similar sizes. For each cross-validation split, we use six groups for training, one group for validation, and one group for testing. The test sets of the eight folds are disjoint and their union is the set of all tasks. 

\paragraph{BinaryClfs}
This dataset contains a wide range of \textbf{binary} \textbf{cl}assi\textbf{f}ication task\textbf{s}, and each task can be described by 1-4 ``yes/no" questions, which we concatenate to the input context as instructions. 
There are in total 204 different tasks, and 73 of them are used for testing, which include sentiment classification, topic classification, definition detection, stance classification, etc. 
We use the same evaluation protocol as in \citet{zhong2021adapting}: 1) we group the tasks by similarity and do not allow training tasks to be similar to testing tasks; 2) we treat ``Yes'' answer as the positive class and calculate the AUC-ROC score for each instruction of each task.

To fit model inputs (concatenation of in-context examples and task input to classify) within the maximum context length (1024) of our LMs, we leave out five evaluation tasks where the maximum task input length exceeds 230 BPE tokens.
We also leave out the spam classification task due to its small test set. 
BinaryClfs does not come with an official validation set. 
To perform hyperparameter tuning, for each testing group, we randomly sample another testing group as its validation group.

\subsection{Implementation Details} \label{sec:details}
\paragraph{Architecture}
We use BERT models for LAMA (BERT-Base [110M parameters], BERT-Large [340M] and DeBERTa-XLarge-V2 [900M]) and GPT2 models for BinaryClfs (GPT2-Medium [345M] and GPT2-Large [774M]). 
We use the Huggingface implementation \cite{wolf2020huggingfaces}.

\paragraph{Hyperparameters}
We select hyperparameters based on few-shot classification accuracy on validation tasks.
Our validation tasks and testing tasks are disjoint, so hyperparameter tuning on validation tasks does not use extra labeled examples on the testing tasks \cite{perez2021true}. 
See Appendix~\ref{sec:appendix-hyperparameters} for the hyperparameters we tuned.

\paragraph{Sampling}
Different instructions and few-shot example choices can lead to different predictions (Section~\ref{sec:ouralgo}). 
At training time, we expose the model to diverse task instructions and few-shot choices by randomly sampling task instructions and few-shot examples for each target example. 

At test time, we report the average accuracy across task instructions and few-shot choices. Since computing the average across all few-shot choices is intractable (there are combinatorically many distinct few-shot choices), we thus calculate the average accuracy of multiple random samplings of few-shot choices as approximation. 
\section{Results}
\label{sec:results}
\begin{table}[t]
\fontsize{9.8}{10}\selectfont
\centering
\begin{tabular}{lcccc}
\toprule
\multirow{2}{*}{} & \multicolumn{2}{c}{\textbf{LAMA}} & \multicolumn{2}{c}{\textbf{BinaryClfs}} \\
\cmidrule(lr){2-3} \cmidrule(lr){4-5}
 & \textbf{BB} & \textbf{BL} & \textbf{GPT2-M} & \textbf{GPT2-L} \\
\midrule
MAML & 16.9 & 21.4 & 63.3 & 63.9 \\
\midrule
ICT & \textbf{19.6} & \textbf{24.3} & \textbf{67.4} & \textbf{69.8}\\
\bottomrule
\end{tabular}
\caption{\label{tab:ict-vs-maml}In-context tuning consistently out-performs MAML on both datasets and all model sizes under the 5-shot setting. BB: BERT-Base. BL: BERT-Large. GPT2-M: GPT2-Medium. GPT2-L: GPT2-Large. }
\end{table}

In-context tuning out-performs MAML and various baselines on the two text classification meta-datasets
(Section \ref{sec:accuracy}). 
It also significantly reduces model sensitivity to instruction wording, example choices, and example ordering compared to prompting raw LMs (Section \ref{sec:sensitivity}). 

\subsection{Few-shot Learning Performance} \label{sec:accuracy}
\paragraph{In-context tuning improves in-context learning accuracy over raw LMs.}
We compare ICT with Raw IC-L in Table~\ref{tab:ict-vs-baselines}. 
In-context tuning consistently out-performs raw LM prompting by 7.6 points on LAMA and 10.6 points on BinaryClfs (averaged across model size and number of few-shots). 
As expected, directly optimizing the few-shot in-context learning objective (Section~\ref{sec:ouralgo}) improves the few-shot in-context learning accuracy.

\paragraph{Few-shot examples lead to more effective task adaptation.} 
We compare few-shot in-context tuning with instruction tuning (equivalent to 0-shot ICT) in Table~\ref{tab:ict-vs-baselines}. Few-shot in-context tuning consistently out-performs instruction tuning on both LAMA and BinaryClfs, with increasing performance gains as number of shots increases. Specifically, we observe that 5-shot in-context tuning out-performs instruction tuning by 6.1 points on LAMA and 4.0 points on BinaryClfs. Results show that demonstration examples besides task instructions facilitate more effective task adaptation.

\paragraph{In-context tuning better leverages the inductive bias for pattern matching.}
By comparing MAML (the first row of Table~\ref{tab:ict-vs-maml}) to instruction tuning (equivalent to 0-shot ICT) of Table~\ref{tab:ict-vs-baselines}, we see that MAML out-performs instruction tuning in most evaluation settings, which indicates that MAML is indeed able to take advantage of the few-shot task examples for task adaptation. However, Table~\ref{tab:ict-vs-maml} shows that our approach of 5-shot in-context tuning out-performs 5-shot MAML consistently on both datasets with an accuracy gain of 2.8 points on LAMA and 5.1 points on BinaryClfs (averaged across model size). We argue that in-context tuning out-performs MAML because in-context tuning better leverages the existing inductive bias of pre-trained LMs to perform pattern matching with in-context examples. 

We also compare in-context tuning to the pipeline of instruction tuning + task-specific fine-tuning (Table~\ref{tab:ict-vs-baselines}). 
Surprisingly, fine-tuning an instruction-tuned model on as few as one task-specific example significantly improves task accuracy, without over-fitting to the few labeled examples. We observe that instruction tuning + 1-shot fine-tuning out-performs instruction tuning (equivalent to 0-shot ICT) by 3.1 points on LAMA (Table~\ref{tab:ict-vs-baselines}). 
Our in-context tuning approach performs comparable or better than instruction tuning + fine-tuning, with increasing accuracy gains as models get bigger (Table~\ref{tab:ict-vs-baselines}). 
For DeBERTa-XLarge-v2 (the largest models we use in this work), in-context tuning out-performs \mbox{InsT + FT}
across all numbers of shots, achieving an accuracy gain of 1.7 points on LAMA (averaged across all numbers of shots). 
We conjecture that in-context tuning will be increasingly effective for bigger models that have a stronger inductive bias of pattern matching. 

\paragraph{In-context tuning reduces the need of task instructions.} 
As coming up with good task instructions can be hard \cite{schick-schutze-2021-exploiting, 10.1162/tacl_a_00324}, we further investigate the effectiveness of in-context tuning without task instructions (Table~\ref{tab:ict-vs-baselines}). In-context tuning is effective in the no-instruction setting as well, consistently out-performing raw in-context learning with no instructions by an average margin of 9.5 points on LAMA. Comparing raw in-context learning with (Raw IC-L) and without instructions (Raw IC-L w/o Ins) \mbox{(Table~\ref{tab:ict-vs-baselines})}, we observe that task instructions yield the most significant performance gains when model size is relatively small (+2.5 points on BERT-Base, +7.7 points on BERT-Large, only +0.6 points on DeBERTa-xlarge). We conjecture that smaller models may be weaker at inferring patterns from in-context examples alone compared to larger models, which is why instructions yield larger performance gains on smaller models. On BERT-Base and BERT-Large models where task instructions are most helpful, in-context tuning reduces the improvement gain from task instructions from 5.1 points (raw in-context learning) to 1.8 points (averaged across BERT-Base and BERT-Large), which indicates that in-context tuning reduces the need of task instructions compared to raw in-context learning. However, we note that instructions still yield performance improvement even if in-context tuning is applied.


\subsection{Sensitivity Analysis}

\label{sec:sensitivity}
We analyze the sensitivity of in-context tuning accuracy with respect to example ordering, example choices, and instruction wording, and compare it with prompting raw LMs. 
Let $I$ denote a random selection of task instruction, $S_T$ a random unordered set of few-shot training examples with size $K$, $\sigma$ a random permutation of $K$ examples.
The accuracy $\mu$ is a function of these three random variables, i.e. $\mu:  (S_T, \sigma, I) \mapsto [0, 1]$. 
We can decompose the total variance of $\mu$ into its variance w.r.t. each of the three random variables, since they are independent (order variance is independent to choice variance because $S_T$ is \emph{unordered}): 

\begin{align*}
    & \quad \text{Var}_{S_T, \sigma, I}[\mu] = \underbrace{\text{Var}_I[\mathbb{E}_{S_T, \sigma}[\mu | I]]}_\text{instruction wording variance} \\
    & + \underbrace{\mathbb{E}_I[\text{Var}_{S_T}[\mathbb{E}_\sigma[\mu | I, S_T]]]}_\text{example choice variance}\\
    &+ \underbrace{\mathbb{E}_{I, S_T}[\text{Var}_{\sigma}[\mu | I, S_T]]}_\text{example order variance}
\end{align*}
We analyze each type of variance below.

\begin{table}[t]
\fontsize{9.8}{10}\selectfont
\centering
\begin{tabular}{lcccc}
\toprule
\multirow{2}{*}{} & \multicolumn{2}{c}{\textbf{LAMA}} & \multicolumn{2}{c}{\textbf{BinaryClfs}} \\
\cmidrule(lr){2-3} \cmidrule(lr){4-5}
& \textbf{BB} & \textbf{BL} & \textbf{GPT2-M} & \textbf{GPT2-L} \\
\midrule
Raw IC-L & 1.82 & 2.14 & 9.26 & 8.84 \\
\midrule
ICT & \textbf{0.66} & \textbf{0.61} & \textbf{1.41} & \textbf{1.58}\\
\bottomrule
\end{tabular}
\caption{\label{tab:sensitivity-ordering}In-context tuning is significantly less sensitive to example ordering compared to in-context learning with raw LMs.}
\end{table}

\begin{table}[t]
\fontsize{9.8}{10}\selectfont
\centering
\begin{tabular}{lcccc}
\toprule
\multirow{2}{*}{} & \multicolumn{2}{c}{\textbf{LAMA}} & \multicolumn{2}{c}{\textbf{BinaryClfs}} \\
\cmidrule(lr){2-3} \cmidrule(lr){4-5}
& \textbf{BB} & \textbf{BL} & \textbf{GPT2-M} & \textbf{GPT2-L} \\
\midrule
Raw IC-L & 3.74 & 6.30 & 18.52 & 20.33 \\
\midrule
ICT & \textbf{1.78} & \textbf{2.57} & \textbf{11.46} & \textbf{11.62} \\
\bottomrule
\end{tabular}
\caption{\label{tab:sensitivity-selection}In-context tuning is significantly less sensitive to example choices compared to in-context learning with raw LMs.}
\end{table}

\begin{table}[t]
\fontsize{9.8}{10}\selectfont
\centering
\begin{tabular}{lcccc}
\toprule
\multirow{2}{*}{} & \multicolumn{2}{c}{BERT-Base} & \multicolumn{2}{c}{BERT-Large} \\
\cmidrule(lr){2-3} \cmidrule(lr){4-5}
& Raw IC-L & ICT & Raw IC-L & ICT \\
\midrule
1-shot & 35.38 & \textbf{26.31} & 34.03 & \textbf{28.78} \\
\midrule
2-shot & 33.79 & \textbf{25.40} & \textbf{17.71} & 19.35 \\
\midrule
5-shot & 24.90 & \textbf{15.64} & 6.36 & \textbf{5.16}\\
\bottomrule
\end{tabular}
\caption{\label{tab:sensitivity-instruction}In-context tuning is much less sensitive to task instruction wording compared to in-context learning with raw LMs.}
\end{table}

\paragraph{In-context tuning is significantly less sensitive to example ordering.} 
We compare the variance with respect to example ordering for in-context tuning and in-context prompting with raw LMs in Table~\ref{tab:sensitivity-ordering}. Results show that in-context tuning is significantly less sensitive to ordering of in-context examples compared to in-context prompting with raw LMs, reducing the sensitivity by 68\% on LAMA and 83\% on BinaryClfs.

\paragraph{In-context tuning is significantly less sensitive to example choices.}
We compare the variance with respect to example choices for in-context tuning and in-context prompting with raw LMs in \mbox{Table~\ref{tab:sensitivity-selection}}.
Results show that in-context tuning is significantly less sensitive to selection of in-context examples compared to in-context prompting with raw LMs across both datasets and all model sizes, reducing the sensitivity by 56\% on LAMA and 40\% on BinaryClfs (averaged across model sizes). We conjecture that in-context tuning is significantly less sensitive to example ordering and selection because the model is exposed to various example orderings and selections during in-context tuning. 

\paragraph{In-context tuning is less sensitive to instruction wording.} 
We report the variance with respect to instruction wording for in-context tuning and in-context prompting with raw LMs in Table~\ref{tab:sensitivity-instruction}.
Results show that in-context tuning is less sensitive to instruction wording compared to in-context prompting with raw LMs in five out of six evaluation settings, reducing the variance by 19\% on LAMA (averaged across model size and number of shots).

We also observe that in-context tuning is especially effective on task instructions with low accuracy under raw in-context learning. For each task, we compute the Pearson correlation between the raw in-context learning accuracy and the accuracy gain from in-context tuning (over raw in-context learning) on all instructions. On the LAMA dataset, we see a strong negative correlation of -0.563 (averaged across all tasks), with p-value $<$ 0.05 on 63\% of the tasks. We conjecture that in-context tuning is much less sensitive to instruction wording because the model is exposed to a wide variety of different task instructions during in-context tuning. 

\paragraph{In-context examples are complementary to instructions.} 
We observe that in-context tuning is especially effective on task instructions with low accuracy under instruction tuning. For each task, we compute the Pearson correlation between the instruction tuning accuracy and the accuracy gain from in-context tuning (over instruction tuning) on all instructions. On the LAMA dataset, we see a strong negative correlation of -0.910 (averaged across all tasks), with p-value $<$ 0.01 on 91\% of the tasks. We conjecture that in-context tuning is much less sensitive to instruction wording because few-shot in-context examples provide additional task information besides the task instructions. 
\section{Related Work}

\paragraph{LM Prompting for FSL}
Pre-trained LMs can be used to perform various FSL tasks when prompted with a natural language task instruction and several task examples \cite{radford2019language, brown2020language, schick-schutze-2021-just, li-liang-2021-prefix, lester2021power, qin-eisner-2021-learning}. However, prompting pre-trained LMs directly for FSL is known to be sensitive to various artifacts, such as the wording of the task instruction and the selection and ordering of few-shot training examples \cite{schick-schutze-2021-exploiting, 10.1162/tacl_a_00324, zhao2021calibrate, gao2021making, liu2021makes}. 
Our work is the first to show that meta-learning with an explicit FSL objective significantly reduces the sensitivity of LM prompting with respect to the in-context examples and instruction wording. 

\paragraph{Meta-learning for FSL}
Meta-learning is a widely used technique in NLP to improve cross-domain transfer \cite{yu-etal-2018-diverse, geng-etal-2019-induction, holla-etal-2020-learning, Deng_2020} and cross-task transfer \cite{gu-etal-2018-meta, bansal-etal-2020-learning, dou-etal-2019-investigating}.
Existing optimization-based meta-learning methods mostly perform task adaptation by fine-tuning a task-agnostic model on task-specific examples using gradient descent \cite{finn2017modelagnostic, jiang2018importance, nichol2018firstorder}. 
However, fine-tuning on few-shot task examples is sensitive to hyperparameters \cite{antoniou2018how} and nested optimization during meta-training is often unstable \cite{nichol2018firstorder, antoniou2018how, NEURIPS2019_072b030b}.
In contrast, our approach performs few-shot task adaptation by using task-specific examples as part of the model input while keeping the model parameters frozen and task-agnostic during the adaptation stage.

\paragraph{Multi-task Learning} In multi-task learning, a single model is trained on the union of training sets of multiple tasks to learn a shared representation \cite{liu-etal-2019-multi}. The multi-task model is then fine-tuned on task-specific examples to adapt to new tasks. Multi-task learning is shown to improve performance on various downstream tasks, especially tasks with small training sets \cite{, khashabi-etal-2020-unifiedqa, ye2021crossfit, aghajanyan2021muppet}. Compared to meta-learning, multi-task learning does not optimize task adaptation directly.

\paragraph{Fine-tuned LMs for Instruction Learning}
Recent work shows that fine-tuning LMs to learn task instructions on a wide variety of tasks can further leverage the inductive bias of LMs to perform instruction learning \cite{zhong2021adapting, mishra2021crosstask, wei2021finetuned}. Our work is partially inspired by this line of work, but we work under the more generic few-shot meta-learning setting, and show that our approach out-performs both instruction tuning and existing few-shot meta-learning methods (e.g., MAML). While previous work focuses on the accuracy improvement gained from instruction fine-tuning, our work also looks into the well-known over-sensitivity issue of FSL and shows that in-context tuning effectively reduces the sensitivity of FSL with respect to various factors. 

Concurrent to our work, \citet{min2021metaicl} also explores in-context tuning under more general Seq2Seq tasks.
In comparison, our work compares in-context tuning to a meta-learning baseline MAML, and shows that in-context tuning mitigates the well-known oversensitivity issue of LM prompting.
Contrary to our paper, \citet{min2021metaicl} finds that in-context tuning under-performs InsT + FT. 
This might be because they use many more shots (16-shot), which could give gradient-based methods more advantage.
\section{Future Directions} \label{sec:future}

\paragraph{Scaling Up and Broader Applications}
Our work only considers simple binary classification and knowledge retrieval tasks, at most 5 in-context examples, and models with fewer than 1 billion parameters.
Nevertheless, it is straightforward to scale up our framework to a wider and more diverse range of general sequence-to-sequence tasks \cite{ye2021crossfit}, more few-shot examples (which requires a longer context size \cite{dai-etal-2019-transformer, wang2020linformer}), and larger models \cite{brown2020language, kaplan2020scaling}.
It is also straightforward to apply in-context tuning to a broader range of scenarios that require adapting to a new setup, e.g., adapting to a new label in classification tasks \cite{xia-etal-2021-incremental}, an unseen database in semantic parsing tasks \cite{suhr-etal-2020-exploring, lee-etal-2021-kaggledbqa}, or a new language pair in machine translation \cite{gu-etal-2018-meta, aharoni-etal-2019-massively}, etc.   

\paragraph{Meta-learning for Robustness} Our work assumed that the few-shot training examples come from the same distribution as the test examples, but this assumption does not necessarily hold in practice. 
For example, the test distribution might constitute new input compositions \cite{lake2018generalization}, rare subgroups \cite{sagawa2019distributionally}, other types of distribution shifts \cite{hendrycks2019benchmarking}, or even adversarial examples \cite{kang2019testing}.
More effective meta-learning methods might learn a more robust learning mechanism and combat these generalization challenges. 

\paragraph{Understanding In-context Learning}
Many properties of in-context learning are still unknown. 
Is in-context learning more robust to distribution shift \cite{lester2021power}?
Can we combine in-context learning and gradient learning to get the benefit of both worlds \cite{wortsman2021robust}? 

\section{Conclusion}
In this work, we propose meta-learning via in-context tuning, which recasts the few-shot learning process of task adaptation and task-specific prediction as a simple sequence prediction problem, where few-shot labeled examples are concatenated with the target example to form the model input. 
In-context tuning out-performs a wide variety of baselines in terms of accuracy, including raw LM prompting, MAML and instruction tuning. 
Meanwhile, sensitivity study shows that our FSL approach of in-context tuning is significantly less sensitive to few-shot examples and instruction wording compared to raw LM prompting.

Given the empirical effectiveness of in-context tuning, we conjecture that the few-shot learning potential of large LMs (e.g., GPT-3) might be broadly underestimated, and that in-context tuning can eliminate well-known artifacts of few-shot LM prompting such as over-sensitivity to example ordering, example selection and instruction wording.

\bibliography{anthology,custom}

\begin{thebibliography}{52}
\expandafter\ifx\csname natexlab\endcsname\relax\def\natexlab#1{#1}\fi

\bibitem[{Adiwardana et~al.(2020)Adiwardana, Luong, So, Hall, Fiedel,
  Thoppilan, Yang, Kulshreshtha, Nemade, Lu, and Le}]{adiwardana2020humanlike}
Daniel Adiwardana, Minh-Thang Luong, David~R. So, Jamie Hall, Noah Fiedel,
  Romal Thoppilan, Zi~Yang, Apoorv Kulshreshtha, Gaurav Nemade, Yifeng Lu, and
  Quoc~V. Le. 2020.
\newblock \href {http://arxiv.org/abs/2001.09977} {Towards a human-like
  open-domain chatbot}.

\bibitem[{Aghajanyan et~al.(2021)Aghajanyan, Gupta, Shrivastava, Chen,
  Zettlemoyer, and Gupta}]{aghajanyan2021muppet}
Armen Aghajanyan, Anchit Gupta, Akshat Shrivastava, Xilun Chen, Luke
  Zettlemoyer, and Sonal Gupta. 2021.
\newblock \href {http://arxiv.org/abs/2101.11038} {Muppet: Massive multi-task
  representations with pre-finetuning}.

\bibitem[{Aharoni et~al.(2019)Aharoni, Johnson, and
  Firat}]{aharoni-etal-2019-massively}
Roee Aharoni, Melvin Johnson, and Orhan Firat. 2019.
\newblock \href {https://doi.org/10.18653/v1/N19-1388} {Massively multilingual
  neural machine translation}.
\newblock In \emph{Proceedings of the 2019 Conference of the North {A}merican
  Chapter of the Association for Computational Linguistics: Human Language
  Technologies, Volume 1 (Long and Short Papers)}, pages 3874--3884,
  Minneapolis, Minnesota. Association for Computational Linguistics.

\bibitem[{Antoniou et~al.(2019)Antoniou, Edwards, and
  Storkey}]{antoniou2018how}
Antreas Antoniou, Harrison Edwards, and Amos Storkey. 2019.
\newblock \href {https://openreview.net/forum?id=HJGven05Y7} {How to train your
  {MAML}}.
\newblock In \emph{International Conference on Learning Representations}.

\bibitem[{Bansal et~al.(2020)Bansal, Jha, and
  McCallum}]{bansal-etal-2020-learning}
Trapit Bansal, Rishikesh Jha, and Andrew McCallum. 2020.
\newblock \href {https://doi.org/10.18653/v1/2020.coling-main.448} {Learning to
  few-shot learn across diverse natural language classification tasks}.
\newblock In \emph{Proceedings of the 28th International Conference on
  Computational Linguistics}, pages 5108--5123, Barcelona, Spain (Online).
  International Committee on Computational Linguistics.

\bibitem[{Brown et~al.(2020)Brown, Mann, Ryder, Subbiah, Kaplan, Dhariwal,
  Neelakantan, Shyam, Sastry, Askell, Agarwal, Herbert-Voss, Krueger, Henighan,
  Child, Ramesh, Ziegler, Wu, Winter, Hesse, Chen, Sigler, Litwin, Gray, Chess,
  Clark, Berner, McCandlish, Radford, Sutskever, and
  Amodei}]{brown2020language}
Tom~B. Brown, Benjamin Mann, Nick Ryder, Melanie Subbiah, Jared Kaplan,
  Prafulla Dhariwal, Arvind Neelakantan, Pranav Shyam, Girish Sastry, Amanda
  Askell, Sandhini Agarwal, Ariel Herbert-Voss, Gretchen Krueger, Tom Henighan,
  Rewon Child, Aditya Ramesh, Daniel~M. Ziegler, Jeffrey Wu, Clemens Winter,
  Christopher Hesse, Mark Chen, Eric Sigler, Mateusz Litwin, Scott Gray,
  Benjamin Chess, Jack Clark, Christopher Berner, Sam McCandlish, Alec Radford,
  Ilya Sutskever, and Dario Amodei. 2020.
\newblock \href {http://arxiv.org/abs/2005.14165} {Language models are few-shot
  learners}.

\bibitem[{Dai et~al.(2019)Dai, Yang, Yang, Carbonell, Le, and
  Salakhutdinov}]{dai-etal-2019-transformer}
Zihang Dai, Zhilin Yang, Yiming Yang, Jaime Carbonell, Quoc Le, and Ruslan
  Salakhutdinov. 2019.
\newblock \href {https://doi.org/10.18653/v1/P19-1285} {Transformer-{XL}:
  Attentive language models beyond a fixed-length context}.
\newblock In \emph{Proceedings of the 57th Annual Meeting of the Association
  for Computational Linguistics}, pages 2978--2988, Florence, Italy.
  Association for Computational Linguistics.

\bibitem[{Deng et~al.(2020)Deng, Zhang, Kang, Zhang, Zhang, and
  Chen}]{Deng_2020}
Shumin Deng, Ningyu Zhang, Jiaojian Kang, Yichi Zhang, Wei Zhang, and Huajun
  Chen. 2020.
\newblock \href {https://doi.org/10.1145/3336191.3371796} {Meta-learning with
  dynamic-memory-based prototypical network for few-shot event detection}.
\newblock \emph{Proceedings of the 13th International Conference on Web Search
  and Data Mining}.

\bibitem[{Dou et~al.(2019)Dou, Yu, and
  Anastasopoulos}]{dou-etal-2019-investigating}
Zi-Yi Dou, Keyi Yu, and Antonios Anastasopoulos. 2019.
\newblock \href {https://doi.org/10.18653/v1/D19-1112} {Investigating
  meta-learning algorithms for low-resource natural language understanding
  tasks}.
\newblock In \emph{Proceedings of the 2019 Conference on Empirical Methods in
  Natural Language Processing and the 9th International Joint Conference on
  Natural Language Processing (EMNLP-IJCNLP)}, pages 1192--1197, Hong Kong,
  China. Association for Computational Linguistics.

\bibitem[{Finn et~al.(2017)Finn, Abbeel, and Levine}]{finn2017modelagnostic}
Chelsea Finn, Pieter Abbeel, and Sergey Levine. 2017.
\newblock \href {http://arxiv.org/abs/1703.03400} {Model-agnostic meta-learning
  for fast adaptation of deep networks}.

\bibitem[{Gao et~al.(2021)Gao, Fisch, and Chen}]{gao2021making}
Tianyu Gao, Adam Fisch, and Danqi Chen. 2021.
\newblock \href {http://arxiv.org/abs/2012.15723} {Making pre-trained language
  models better few-shot learners}.

\bibitem[{Geng et~al.(2019)Geng, Li, Li, Zhu, Jian, and
  Sun}]{geng-etal-2019-induction}
Ruiying Geng, Binhua Li, Yongbin Li, Xiaodan Zhu, Ping Jian, and Jian Sun.
  2019.
\newblock \href {https://doi.org/10.18653/v1/D19-1403} {Induction networks for
  few-shot text classification}.
\newblock In \emph{Proceedings of the 2019 Conference on Empirical Methods in
  Natural Language Processing and the 9th International Joint Conference on
  Natural Language Processing (EMNLP-IJCNLP)}, pages 3904--3913, Hong Kong,
  China. Association for Computational Linguistics.

\bibitem[{Gu et~al.(2018)Gu, Wang, Chen, Li, and Cho}]{gu-etal-2018-meta}
Jiatao Gu, Yong Wang, Yun Chen, Victor O.~K. Li, and Kyunghyun Cho. 2018.
\newblock \href {https://doi.org/10.18653/v1/D18-1398} {Meta-learning for
  low-resource neural machine translation}.
\newblock In \emph{Proceedings of the 2018 Conference on Empirical Methods in
  Natural Language Processing}, pages 3622--3631, Brussels, Belgium.
  Association for Computational Linguistics.

\bibitem[{He et~al.(2016)He, Zhang, Ren, and Sun}]{7780459}
Kaiming He, Xiangyu Zhang, Shaoqing Ren, and Jian Sun. 2016.
\newblock \href {https://doi.org/10.1109/CVPR.2016.90} {Deep residual learning
  for image recognition}.
\newblock In \emph{2016 IEEE Conference on Computer Vision and Pattern
  Recognition (CVPR)}, pages 770--778.

\bibitem[{Hendrycks and Dietterich(2019)}]{hendrycks2019benchmarking}
Dan Hendrycks and Thomas Dietterich. 2019.
\newblock Benchmarking neural network robustness to common corruptions and
  perturbations.
\newblock \emph{arXiv preprint arXiv:1903.12261}.

\bibitem[{Holla et~al.(2020)Holla, Mishra, Yannakoudakis, and
  Shutova}]{holla-etal-2020-learning}
Nithin Holla, Pushkar Mishra, Helen Yannakoudakis, and Ekaterina Shutova. 2020.
\newblock \href {https://doi.org/10.18653/v1/2020.findings-emnlp.405} {Learning
  to learn to disambiguate: {M}eta-learning for few-shot word sense
  disambiguation}.
\newblock In \emph{Findings of the Association for Computational Linguistics:
  EMNLP 2020}, pages 4517--4533, Online. Association for Computational
  Linguistics.

\bibitem[{Jiang et~al.(2019)Jiang, Havaei, Chartrand, Chouaib, Vincent, Jesson,
  Chapados, and Matwin}]{jiang2018importance}
Xiang Jiang, Mohammad Havaei, Gabriel Chartrand, Hassan Chouaib, Thomas
  Vincent, Andrew Jesson, Nicolas Chapados, and Stan Matwin. 2019.
\newblock \href {https://openreview.net/forum?id=SyxMWh09KX} {Attentive
  task-agnostic meta-learning for few-shot text classification}.

\bibitem[{Jiang et~al.(2020)Jiang, Xu, Araki, and
  Neubig}]{10.1162/tacl_a_00324}
Zhengbao Jiang, Frank~F. Xu, Jun Araki, and Graham Neubig. 2020.
\newblock \href {https://doi.org/10.1162/tacl_a_00324} {{How Can We Know What
  Language Models Know?}}
\newblock \emph{Transactions of the Association for Computational Linguistics},
  8:423--438.

\bibitem[{Kang et~al.(2019)Kang, Sun, Hendrycks, Brown, and
  Steinhardt}]{kang2019testing}
Daniel Kang, Yi~Sun, Dan Hendrycks, Tom Brown, and Jacob Steinhardt. 2019.
\newblock Testing robustness against unforeseen adversaries.
\newblock \emph{arXiv preprint arXiv:1908.08016}.

\bibitem[{Kaplan et~al.(2020)Kaplan, McCandlish, Henighan, Brown, Chess, Child,
  Gray, Radford, Wu, and Amodei}]{kaplan2020scaling}
Jared Kaplan, Sam McCandlish, Tom Henighan, Tom~B Brown, Benjamin Chess, Rewon
  Child, Scott Gray, Alec Radford, Jeffrey Wu, and Dario Amodei. 2020.
\newblock Scaling laws for neural language models.
\newblock \emph{arXiv preprint arXiv:2001.08361}.

\bibitem[{Khashabi et~al.(2020)Khashabi, Min, Khot, Sabharwal, Tafjord, Clark,
  and Hajishirzi}]{khashabi-etal-2020-unifiedqa}
Daniel Khashabi, Sewon Min, Tushar Khot, Ashish Sabharwal, Oyvind Tafjord,
  Peter Clark, and Hannaneh Hajishirzi. 2020.
\newblock \href {https://doi.org/10.18653/v1/2020.findings-emnlp.171}
  {{UNIFIEDQA}: Crossing format boundaries with a single {QA} system}.
\newblock In \emph{Findings of the Association for Computational Linguistics:
  EMNLP 2020}, pages 1896--1907, Online. Association for Computational
  Linguistics.

\bibitem[{Lake and Baroni(2018)}]{lake2018generalization}
Brenden Lake and Marco Baroni. 2018.
\newblock Generalization without systematicity: On the compositional skills of
  sequence-to-sequence recurrent networks.
\newblock In \emph{International conference on machine learning}, pages
  2873--2882. PMLR.

\bibitem[{Lake et~al.(2016)Lake, Ullman, Tenenbaum, and
  Gershman}]{lake2016building}
Brenden~M. Lake, Tomer~D. Ullman, Joshua~B. Tenenbaum, and Samuel~J. Gershman.
  2016.
\newblock \href {http://arxiv.org/abs/1604.00289} {Building machines that learn
  and think like people}.

\bibitem[{Lee et~al.(2021)Lee, Polozov, and
  Richardson}]{lee-etal-2021-kaggledbqa}
Chia-Hsuan Lee, Oleksandr Polozov, and Matthew Richardson. 2021.
\newblock \href {https://doi.org/10.18653/v1/2021.acl-long.176}
  {{K}aggle{DBQA}: Realistic evaluation of text-to-{SQL} parsers}.
\newblock In \emph{Proceedings of the 59th Annual Meeting of the Association
  for Computational Linguistics and the 11th International Joint Conference on
  Natural Language Processing (Volume 1: Long Papers)}, pages 2261--2273,
  Online. Association for Computational Linguistics.

\bibitem[{Lester et~al.(2021)Lester, Al-Rfou, and Constant}]{lester2021power}
Brian Lester, Rami Al-Rfou, and Noah Constant. 2021.
\newblock The power of scale for parameter-efficient prompt tuning.
\newblock \emph{arXiv preprint arXiv:2104.08691}.

\bibitem[{Li and Liang(2021)}]{li-liang-2021-prefix}
Xiang~Lisa Li and Percy Liang. 2021.
\newblock \href {https://doi.org/10.18653/v1/2021.acl-long.353} {Prefix-tuning:
  Optimizing continuous prompts for generation}.
\newblock In \emph{Proceedings of the 59th Annual Meeting of the Association
  for Computational Linguistics and the 11th International Joint Conference on
  Natural Language Processing (Volume 1: Long Papers)}, pages 4582--4597,
  Online. Association for Computational Linguistics.

\bibitem[{Liu et~al.(2021)Liu, Shen, Zhang, Dolan, Carin, and
  Chen}]{liu2021makes}
Jiachang Liu, Dinghan Shen, Yizhe Zhang, Bill Dolan, Lawrence Carin, and Weizhu
  Chen. 2021.
\newblock \href {http://arxiv.org/abs/2101.06804} {What makes good in-context
  examples for gpt-$3$?}

\bibitem[{Liu et~al.(2019)Liu, He, Chen, and Gao}]{liu-etal-2019-multi}
Xiaodong Liu, Pengcheng He, Weizhu Chen, and Jianfeng Gao. 2019.
\newblock \href {https://doi.org/10.18653/v1/P19-1441} {Multi-task deep neural
  networks for natural language understanding}.
\newblock In \emph{Proceedings of the 57th Annual Meeting of the Association
  for Computational Linguistics}, pages 4487--4496, Florence, Italy.
  Association for Computational Linguistics.

\bibitem[{Min et~al.(2021)Min, Lewis, Zettlemoyer, and
  Hajishirzi}]{min2021metaicl}
Sewon Min, Mike Lewis, Luke Zettlemoyer, and Hannaneh Hajishirzi. 2021.
\newblock Metaicl: Learning to learn in context.
\newblock \emph{arXiv preprint arXiv:2110.15943}.

\bibitem[{Mishra et~al.(2021)Mishra, Khashabi, Baral, and
  Hajishirzi}]{mishra2021crosstask}
Swaroop Mishra, Daniel Khashabi, Chitta Baral, and Hannaneh Hajishirzi. 2021.
\newblock \href {http://arxiv.org/abs/2104.08773} {Cross-task generalization
  via natural language crowdsourcing instructions}.

\bibitem[{Nichol et~al.(2018)Nichol, Achiam, and
  Schulman}]{nichol2018firstorder}
Alex Nichol, Joshua Achiam, and John Schulman. 2018.
\newblock \href {http://arxiv.org/abs/1803.02999} {On first-order meta-learning
  algorithms}.

\bibitem[{Nikolaev et~al.(2020)Nikolaev, Arviv, Karidi, Kenneth, Mitnik,
  Saeboe, and Abend}]{nikolaev-etal-2020-fine}
Dmitry Nikolaev, Ofir Arviv, Taelin Karidi, Neta Kenneth, Veronika Mitnik,
  Lilja~Maria Saeboe, and Omri Abend. 2020.
\newblock \href {https://doi.org/10.18653/v1/2020.acl-main.109} {Fine-grained
  analysis of cross-linguistic syntactic divergences}.
\newblock In \emph{Proceedings of the 58th Annual Meeting of the Association
  for Computational Linguistics}, pages 1159--1176, Online. Association for
  Computational Linguistics.

\bibitem[{Perez et~al.(2021)Perez, Kiela, and Cho}]{perez2021true}
Ethan Perez, Douwe Kiela, and Kyunghyun Cho. 2021.
\newblock \href {http://arxiv.org/abs/2105.11447} {True few-shot learning with
  language models}.

\bibitem[{Petroni et~al.(2019)Petroni, Rockt{\"a}schel, Riedel, Lewis, Bakhtin,
  Wu, and Miller}]{petroni-etal-2019-language}
Fabio Petroni, Tim Rockt{\"a}schel, Sebastian Riedel, Patrick Lewis, Anton
  Bakhtin, Yuxiang Wu, and Alexander Miller. 2019.
\newblock \href {https://doi.org/10.18653/v1/D19-1250} {Language models as
  knowledge bases?}
\newblock In \emph{Proceedings of the 2019 Conference on Empirical Methods in
  Natural Language Processing and the 9th International Joint Conference on
  Natural Language Processing (EMNLP-IJCNLP)}, pages 2463--2473, Hong Kong,
  China. Association for Computational Linguistics.

\bibitem[{Poerner et~al.(2020)Poerner, Waltinger, and
  Sch{\"u}tze}]{poerner-etal-2020-e}
Nina Poerner, Ulli Waltinger, and Hinrich Sch{\"u}tze. 2020.
\newblock \href {https://doi.org/10.18653/v1/2020.findings-emnlp.71}
  {{E}-{BERT}: Efficient-yet-effective entity embeddings for {BERT}}.
\newblock In \emph{Findings of the Association for Computational Linguistics:
  EMNLP 2020}, pages 803--818, Online. Association for Computational
  Linguistics.

\bibitem[{Qin and Eisner(2021)}]{qin-eisner-2021-learning}
Guanghui Qin and Jason Eisner. 2021.
\newblock \href {https://doi.org/10.18653/v1/2021.naacl-main.410} {Learning how
  to ask: Querying {LM}s with mixtures of soft prompts}.
\newblock In \emph{Proceedings of the 2021 Conference of the North American
  Chapter of the Association for Computational Linguistics: Human Language
  Technologies}, pages 5203--5212, Online. Association for Computational
  Linguistics.

\bibitem[{Radford et~al.(2019)Radford, Wu, Child, Luan, Amodei, and
  Sutskever}]{radford2019language}
Alec Radford, Jeff Wu, Rewon Child, David Luan, Dario Amodei, and Ilya
  Sutskever. 2019.
\newblock Language models are unsupervised multitask learners.

\bibitem[{Rajeswaran et~al.(2019)Rajeswaran, Finn, Kakade, and
  Levine}]{NEURIPS2019_072b030b}
Aravind Rajeswaran, Chelsea Finn, Sham~M Kakade, and Sergey Levine. 2019.
\newblock \href
  {https://proceedings.neurips.cc/paper/2019/file/072b030ba126b2f4b2374f342be9ed44-Paper.pdf}
  {Meta-learning with implicit gradients}.
\newblock In \emph{Advances in Neural Information Processing Systems},
  volume~32. Curran Associates, Inc.

\bibitem[{Sagawa et~al.(2019)Sagawa, Koh, Hashimoto, and
  Liang}]{sagawa2019distributionally}
Shiori Sagawa, Pang~Wei Koh, Tatsunori~B Hashimoto, and Percy Liang. 2019.
\newblock Distributionally robust neural networks for group shifts: On the
  importance of regularization for worst-case generalization.
\newblock \emph{arXiv preprint arXiv:1911.08731}.

\bibitem[{Schick and
  Sch{\"u}tze(2021{\natexlab{a}})}]{schick-schutze-2021-exploiting}
Timo Schick and Hinrich Sch{\"u}tze. 2021{\natexlab{a}}.
\newblock \href {https://aclanthology.org/2021.eacl-main.20} {Exploiting
  cloze-questions for few-shot text classification and natural language
  inference}.
\newblock In \emph{Proceedings of the 16th Conference of the European Chapter
  of the Association for Computational Linguistics: Main Volume}, pages
  255--269, Online. Association for Computational Linguistics.

\bibitem[{Schick and
  Sch{\"u}tze(2021{\natexlab{b}})}]{schick-schutze-2021-just}
Timo Schick and Hinrich Sch{\"u}tze. 2021{\natexlab{b}}.
\newblock \href {https://doi.org/10.18653/v1/2021.naacl-main.185} {It{'}s not
  just size that matters: Small language models are also few-shot learners}.
\newblock In \emph{Proceedings of the 2021 Conference of the North American
  Chapter of the Association for Computational Linguistics: Human Language
  Technologies}, pages 2339--2352, Online. Association for Computational
  Linguistics.

\bibitem[{Silver et~al.(2016)Silver, Huang, Maddison, Guez, Sifre, Driessche,
  Schrittwieser, Antonoglou, Panneershelvam, Lanctot, Dieleman, Grewe, Nham,
  Kalchbrenner, Sutskever, Lillicrap, Leach, Kavukcuoglu, Graepel, and
  Hassabis}]{alphago}
David Silver, Aja Huang, Christopher Maddison, Arthur Guez, Laurent Sifre,
  George Driessche, Julian Schrittwieser, Ioannis Antonoglou, Veda
  Panneershelvam, Marc Lanctot, Sander Dieleman, Dominik Grewe, John Nham, Nal
  Kalchbrenner, Ilya Sutskever, Timothy Lillicrap, Madeleine Leach, Koray
  Kavukcuoglu, Thore Graepel, and Demis Hassabis. 2016.
\newblock \href {https://doi.org/10.1038/nature16961} {Mastering the game of go
  with deep neural networks and tree search}.
\newblock \emph{Nature}, 529:484--489.

\bibitem[{Suhr et~al.(2020)Suhr, Chang, Shaw, and
  Lee}]{suhr-etal-2020-exploring}
Alane Suhr, Ming-Wei Chang, Peter Shaw, and Kenton Lee. 2020.
\newblock \href {https://doi.org/10.18653/v1/2020.acl-main.742} {Exploring
  unexplored generalization challenges for cross-database semantic parsing}.
\newblock In \emph{Proceedings of the 58th Annual Meeting of the Association
  for Computational Linguistics}, pages 8372--8388, Online. Association for
  Computational Linguistics.

\bibitem[{Wang et~al.(2020)Wang, Li, Khabsa, Fang, and Ma}]{wang2020linformer}
Sinong Wang, Belinda~Z. Li, Madian Khabsa, Han Fang, and Hao Ma. 2020.
\newblock \href {http://arxiv.org/abs/2006.04768} {Linformer: Self-attention
  with linear complexity}.

\bibitem[{Wei et~al.(2021)Wei, Bosma, Zhao, Guu, Yu, Lester, Du, Dai, and
  Le}]{wei2021finetuned}
Jason Wei, Maarten Bosma, Vincent~Y. Zhao, Kelvin Guu, Adams~Wei Yu, Brian
  Lester, Nan Du, Andrew~M. Dai, and Quoc~V. Le. 2021.
\newblock \href {http://arxiv.org/abs/2109.01652} {Finetuned language models
  are zero-shot learners}.

\bibitem[{Wolf et~al.(2020)Wolf, Debut, Sanh, Chaumond, Delangue, Moi, Cistac,
  Rault, Louf, Funtowicz, Davison, Shleifer, von Platen, Ma, Jernite, Plu, Xu,
  Scao, Gugger, Drame, Lhoest, and Rush}]{wolf2020huggingfaces}
Thomas Wolf, Lysandre Debut, Victor Sanh, Julien Chaumond, Clement Delangue,
  Anthony Moi, Pierric Cistac, Tim Rault, Rémi Louf, Morgan Funtowicz, Joe
  Davison, Sam Shleifer, Patrick von Platen, Clara Ma, Yacine Jernite, Julien
  Plu, Canwen Xu, Teven~Le Scao, Sylvain Gugger, Mariama Drame, Quentin Lhoest,
  and Alexander~M. Rush. 2020.
\newblock \href {http://arxiv.org/abs/1910.03771} {Huggingface's transformers:
  State-of-the-art natural language processing}.

\bibitem[{Wortsman et~al.(2021)Wortsman, Ilharco, Li, Kim, Hajishirzi, Farhadi,
  Namkoong, and Schmidt}]{wortsman2021robust}
Mitchell Wortsman, Gabriel Ilharco, Mike Li, Jong~Wook Kim, Hannaneh
  Hajishirzi, Ali Farhadi, Hongseok Namkoong, and Ludwig Schmidt. 2021.
\newblock Robust fine-tuning of zero-shot models.
\newblock \emph{arXiv preprint arXiv:2109.01903}.

\bibitem[{Xia et~al.(2021)Xia, Yin, Feng, and Yu}]{xia-etal-2021-incremental}
Congying Xia, Wenpeng Yin, Yihao Feng, and Philip Yu. 2021.
\newblock \href {https://doi.org/10.18653/v1/2021.naacl-main.106} {Incremental
  few-shot text classification with multi-round new classes: Formulation,
  dataset and system}.
\newblock In \emph{Proceedings of the 2021 Conference of the North American
  Chapter of the Association for Computational Linguistics: Human Language
  Technologies}, pages 1351--1360, Online. Association for Computational
  Linguistics.

\bibitem[{Ye et~al.(2021)Ye, Lin, and Ren}]{ye2021crossfit}
Qinyuan Ye, Bill~Yuchen Lin, and Xiang Ren. 2021.
\newblock Crossfit: A few-shot learning challenge for cross-task generalization
  in nlp.
\newblock \emph{arXiv preprint arXiv:2104.08835}.

\bibitem[{Yu et~al.(2018)Yu, Guo, Yi, Chang, Potdar, Cheng, Tesauro, Wang, and
  Zhou}]{yu-etal-2018-diverse}
Mo~Yu, Xiaoxiao Guo, Jinfeng Yi, Shiyu Chang, Saloni Potdar, Yu~Cheng, Gerald
  Tesauro, Haoyu Wang, and Bowen Zhou. 2018.
\newblock \href {https://doi.org/10.18653/v1/N18-1109} {Diverse few-shot text
  classification with multiple metrics}.
\newblock In \emph{Proceedings of the 2018 Conference of the North {A}merican
  Chapter of the Association for Computational Linguistics: Human Language
  Technologies, Volume 1 (Long Papers)}, pages 1206--1215, New Orleans,
  Louisiana. Association for Computational Linguistics.

\bibitem[{Zhao et~al.(2021)Zhao, Wallace, Feng, Klein, and
  Singh}]{zhao2021calibrate}
Tony~Z. Zhao, Eric Wallace, Shi Feng, Dan Klein, and Sameer Singh. 2021.
\newblock \href {http://arxiv.org/abs/2102.09690} {Calibrate before use:
  Improving few-shot performance of language models}.

\bibitem[{Zhong et~al.(2021)Zhong, Lee, Zhang, and Klein}]{zhong2021adapting}
Ruiqi Zhong, Kristy Lee, Zheng Zhang, and Dan Klein. 2021.
\newblock \href {http://arxiv.org/abs/2104.04670} {Adapting language models for
  zero-shot learning by meta-tuning on dataset and prompt collections}.

\end{thebibliography}
\bibliographystyle{acl_natbib}

\pagebreak

\appendix

\section{Hyperparameters}
\label{sec:appendix-hyperparameters}
In this section, we report the hyperparameters we tuned for our approach and each baseline. 

\paragraph{In-Context Tuning (ours)} We tune number of training epochs ([10, 15, 30] for LAMA and [1e-7, 3e-7, 1e-6, 3e-6] for BinaryClfs) and learning rate ([1e-7, 3e-7, 1e-6, 3e-6] for LAMA and [3e-6, 1e-5, 3e-5, 1e-4] for BinaryClfs). 

\paragraph{MAML} We assume that inner optimization and outer optimization use the same learning rate. We tuned number of adapt steps ([1, 2, 4] for both datasets) and learning rate ([3e-7, 1e-6, 3e-6, 1e-5, 3e-5, 1e-4, 3e-4, 1e-3] for LAMA and [3e-6, 1e-5, 3e-5, 1e-4, 3e-4, 1e-3] for BinaryClfs).

\paragraph{Instruction-Tuning + Fine-tuning} For instruction tuning we tuned the same set of hyperparameters as in in-context tuning. The instruction tuning model with the highest validation performance are used for downstream task fine-tuning. For task fine-tuning, we tuned number of training epochs ([5, 10, 15, 30, 40] for LAMA and [5, 10, 15, 30, 40] for BinaryClfs) and learning rate ([1e-7, 3e-7, 1e-6, 3e-6, 1e-5, 3e-5] for LAMA and [3e-6, 1e-5, 3e-5, 1e-4, 3e-4, 1e-3] for BinaryClfs).


\end{document}